\documentclass[letterpaper, 10 pt, conference]{Resources/IEEEconf}  

\IEEEoverridecommandlockouts                              

\usepackage{etoolbox}
\makeatletter
\patchcmd{\@makecaption}
  {\scshape}
  {}
  {}
  {}
\makeatletter
\patchcmd{\@makecaption}
  {\\}
  {.\ }
  {}
  {}
\makeatother

\usepackage{gensymb}
\usepackage{graphicx}
\let\labelindent\relax 
\usepackage{lipsum}  
\usepackage{multirow}
\usepackage{hyperref}
\usepackage{enumitem}
\usepackage{xcolor}
\usepackage{array}
\newcolumntype{x}[1]{>{\centering\let\newline\\\arraybackslash\hspace{0pt}}p{#1}} 
\usepackage{svg}
\usepackage{amsmath}
\svgpath{{../Figures/}}
\usepackage{float} 
\usepackage{algorithm}
\usepackage{wrapfig}
\usepackage[noend]{algpseudocode}
\usepackage{lipsum}

\title{\LARGE \bf
Contact-Free Grasp Stability Prediction with In-Hand Time-of-Flight Sensors
}
\author{
Kyle DuFrene\textsuperscript{1}, Cindy Grimm\textsuperscript{1}
\thanks{\textsuperscript{1}Collaborative Robotics and Intelligent Systems (CoRIS) Institute, Oregon State University, Corvallis, OR 97331 
\newline\indent This work was supported in part by the NSF under Grants CCRI 1925715 and RI 1911050.}}

\begin{document}
\maketitle
\thispagestyle{empty}
\pagestyle{empty}

\begin{abstract}
Current approaches to grasp planning for robotics demonstrate high success rates, but degrade with noisy sensors and other factors. Previous works have proposed tactile-based grasp stability classifiers to detect failures, but these approaches rely on making contact and grasping the object to do so. We propose a contact-free grasp stability predictor using multi-zone time-of-flight sensors mounted in the distal links of a gripper. Our method, as it does not require grasping the object to make a prediction, significantly speeds up the stability classification process, cycling at 15 Hz. We collected over 2,500 real-world grasps across 15 objects to train a classifier. Additionally, we conducted grasp attempts over six additional unseen objects, three for validation and model selection, and three for model testing. Our approach demonstrated strong classification performance, with an accuracy of 85.5\% on validation and 86.0\% on test objects. 
\end{abstract}

\section{Introduction}
Robotic grasping is important as it allows robots to interact within human environments or operate in domains deemed unsafe or inaccessible to humans. Within the variety of these complex environments -- home robotics, production assembly environments, space -- robust grasping capabilities are a must for commercial viability. Necessary components in a robotic grasping pipeline include perception and grasp prediction algorithms, but due to noisy sensor data, cluttered environments, and generalized algorithms, only 70-95\% of grasps attempted by current grasp prediction algorithms are stable~\cite{google, science, pick_place}. We propose a novel method to predict grasp stability {\em just before grasping} using in-finger time-of-flight sensors. This method can be used in conjunction with current grasping pipelines to evaluate if the suggested grasp is likely to succeed during the actual grasp.

Typically, when a grasp attempt fails on execution, the object is displaced or dropped partially through a movement. Because the object has moved, the grasping process must start over again to generate a new set of potential grasps. This is not always feasible for fragile, breakable objects or in time-constrained situations. To improve robustness while executing the grasp, various works have measured grasp stability {\em after} contact using tactile sensors~\cite{stability_2011,feeling_classifier, tactile_regrasp, tactile_approach}. For example, in~\cite{feeling_classifier} the authors correctly predict grasp stability using only tactile sensors in 75\% of trials. Although these methods reduce the chance of dropping the object after grasping, tactile approaches can still result in objects being knocked over.

Unlike tactile approaches (which require contact with the object), other works have attempted to classify grasps at the near-contact stage (after positioning the hand but before closing the fingers). Swenson et al. proposed a method using multiple in-hand ranging sensors and information about object geometry and pose~\cite{nigel}. The authors saw good results --- 90\% classification accuracy with limited sensor noise. However, their approach requires ground-truth information about object geometry and pose, which is infeasible in many grasping situations. 

We propose a generalizable approach to predicting grasp success without making contact with the object. After moving to a grasp pose, but before closing the gripper, distance data is recorded from a multi-zone time-of-flight sensor mounted in each distal link. We combine this data with finger joint angles to predict the success of the grasp using a trained random forest network. The data collection is fast and can be integrated into existing grasping pipelines at 15 Hz, enabling real-world deployment. 

\begin{figure}[t!]
    \centering
    \includegraphics[width=0.48\textwidth]{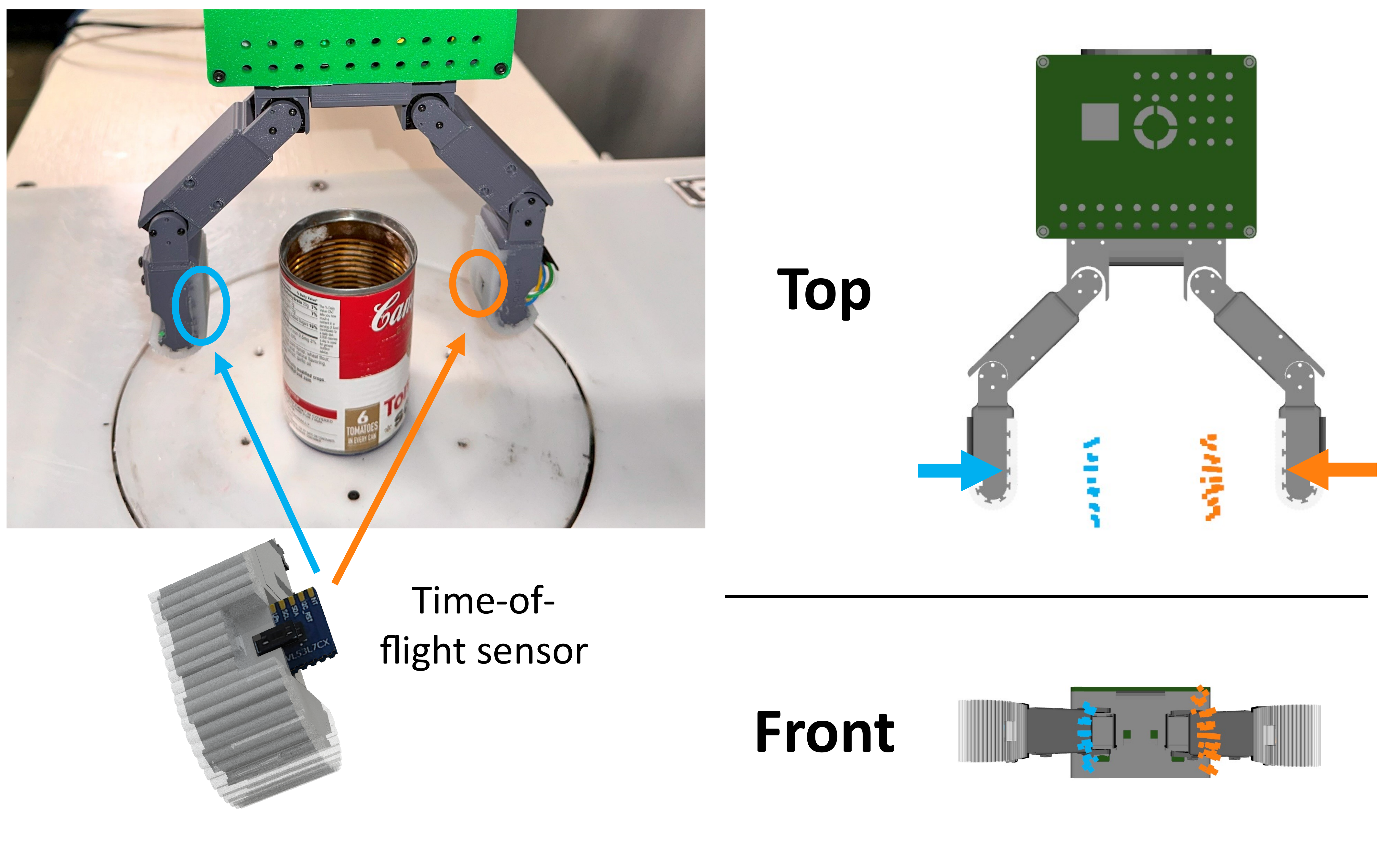}
    \caption{\textbf{\textit{Left}} - The custom gripper mounted to the end of a Kinova Gen 3 arm. The soup can from the YCB object set is placed between the fingers. The location of the time-of-flight sensors is marked in blue and orange. A cutaway view of the distal link exposing the TOF sensor is shown below. \textbf{\textit{Right}} - Top and side views from RVIZ showing the TOF sensor data. Orange points are from the left TOF sensor, blue from the right TOF sensor.}
    \label{fig:tof_visual}
\end{figure}

Our approach predicts the grasp outcome \textit{before} fingers close, which has three unique benefits. First, unlike metrics of grasp success that only monitor gripper position or tactile sensor data, no contact is required and the target object is not disturbed (i.e. knocked over or shifted). Second, our approach requires no prior knowledge of object geometry. Lastly, our approach enables active pose modification. For example, if a grasp is predicted as a failure, one could move the gripper to the next best predicted grasp pose or servo the gripper small distances until a successful grasp is predicted.

We train our predictive random forest network with 1,990 \emph{real-world} grasp attempts across 15 objects using an automated Grasp Reset Mechanism~\cite{dufrene2024grasp}. For each trial, we attempt to grasp the object and move it to a drop-off area. In this work, a grasp is determined successful if and only if the object ends in the drop-off area (the gripper successfully grasped and \textit{maintained} the grasp on the object). Training grasp attempts includes a mixture of discrete poses sampled around the object. The collection of objects is comprised of geometric shapes (such as rectangular prisms and cylinders) and a variety of objects from the YCB object set~\cite{YCBDataset2015}. The selected model achieves 85.5\% prediction accuracy on three unseen validation objects and 86.0\% on three additional unseen test objects.

Our contributions in this work include:
\begin{itemize}
    \item A contact-free predictive approach to determining grasp stability prior to the grasp attempt, with no prior knowledge of the object required. 
    \item Validation of our approach with physical trials on unseen objects.
    \item A demonstration of the usability of low-cost time-of-flight sensors for robust grasping.
\end{itemize}

\section{Related Work}
Predicting grasp stability is an integral part of existing grasp planners. Early methods used analytical techniques, such as form-closure with object models~\cite{overview}. Numerous works have demonstrated the ability to predict grasp success with these geometric, or kinematic or dynamic, techniques~\cite{quality_pose_uncertainty, grasp_metrics, grasp_metric_2,geometric}. However, these methods require ground-truth information or make assumptions about object properties that are impractical in real-world scenarios~\cite{data_driven}. Recent approaches have shifted to data-driven methods, commonly using deep learning~\cite{deep_learning} on images or point clouds taken from cameras either on the robot arm or an external viewpoint. Many of these methods do not require ground truth information about the object shape and achieve grasp success rates of 70-95\%~\cite{google,science,pick_place}.

To improve accuracy, various authors have proposed using tactile sensors to predict grasp stability after the gripper is closed~\cite{stability_2011, feeling_classifier, tactile_regrasp, tactile_approach, stability}. For example, Calandra et al. trained a deep neural network to predict grasp success using only tactile data from GelSight sensors and saw a 75\% accuracy in predicting grasp outcome~\cite{feeling_classifier}. Similarly, Zhao et al. proposed an approach using Mini-TacTip sensors with a novel convolutional layer and achieved almost 93\% accuracy across three test objects~\cite{tactile_approach}. Calandra et al. then expanded on their previous work by using tactile and visual data to predict grasp success of a future re-grasp --- predicting the optimal new grasp within a position, orientation, and force modification range~\cite{tactile_regrasp}. Other multimodal approaches combine tactile data with additional modalities such as external cameras or object mass, yielding up to 95\% stability-classification accuracy~\cite{multimodal}, and slip-detection methods combining tactile sensing with wrist-mounted cameras report 88\% accuracy~\cite{slip}. While effective, these methods rely on deep neural networks that require large training datasets and substantial computational resources.



To reduce computational complexity, Li et al. proposed an approach using a low resolution tactile sensor~\cite{tactile_simple}. Although they reported a 98\% classification rate, the approach was only evaluated on six objects and was not tested on unseen objects, limiting conclusions about generalization.

Despite strong performance, all tactile methods require physical contact with the object, which can lead to undesirable interactions such as knocking over delicate objects, and potentially adds latency if the object is disturbed and the perception/planning pipeline must start over.

In contrasts, Swenson et al. proposed a grasp stability classification method using in-hand rangefinders and object information (size and position)~\cite{nigel}. The author's approach has the benefit of predicting grasp outcome with \emph{no contact}, and demonstrated an accuracy of 96\% in simulation. However, accuracy declined to a maximum of 76\% in real-world tests and the method requires reasonably accurate ground-truth object information (<7.5\% noise) and a large number of rangefinders distributed across the hand. 

Sasaki et al. similarly employed multiple proximity sensors to update the pose of a gripper~\cite{8594430}. Although effective when correcting against support surfaces, the work provides limited evidence of improved grasp success.

Our proposed method addresses these shortcomings by using only a single time-of-flight sensor in each distal link and requiring no {\em a priori} object information.

Beyond grasping and manipulation, recent work has demonstrated that multi-zone TOF sensors provide insightful data for robotics. For example, multi-zone TOF sensors have been used for human presence and attention detection \cite{presence}. Similar sensors were also integrated on robot arms for proximity detection \cite{proxyskin}.

\section{Methods}
\begin{figure*}
    \centering
    \includegraphics[width=1\textwidth]{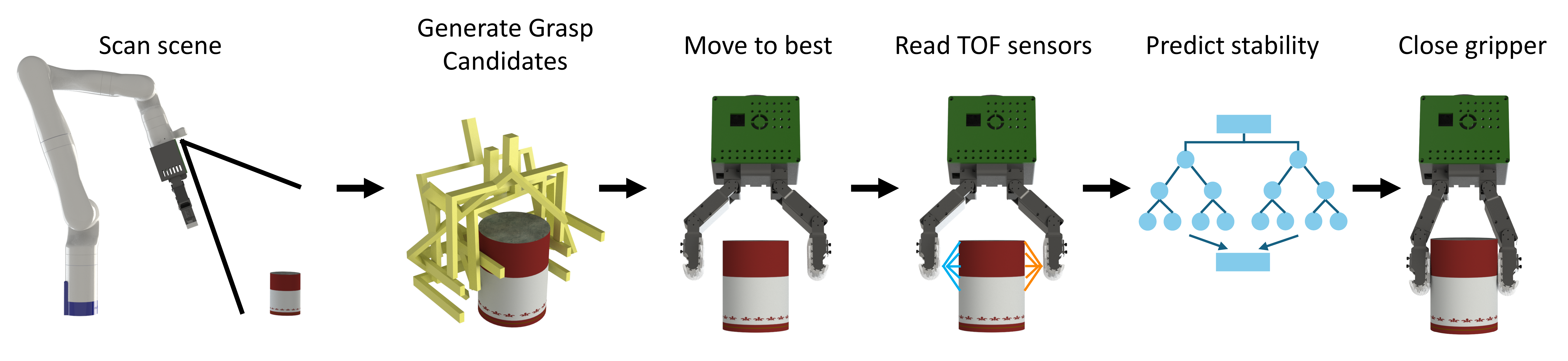}
    \caption{A diagram of how our grasp stability classification would fit into existing grasping pipelines. Our approach does not require contact with the object to predict grasp stability.}
    \label{fig:flowchart}
\end{figure*}
In this work, we present a contact-free approach to predict grasp stability before closing the gripper on an object. This is achieved using in-hand time-of-flight sensors with a machine learning classifier. Our approach enables robust grasping without knowledge of the object (assuming the object is graspable). All work within this paper is performed in the real-world including data collection, training, and testing.

We first present our prediction approach in Section~\ref{predict}. Then, we show how our approach can improve a grasping pipeline in Section~\ref{pipeline}. Data collection and classifier training is discussed in Section~\ref{train}. Finally, we present the custom gripper and real-world hardware in Section~\ref{hardware2}.

\subsection{Predictive Approach (hardware)}\label{predict}
Our predictive approach uses two multi-zone time-of-flight (TOF) sensors, one mounted in each distal link, with a trained classifier to predict grasp outcome. We used a fully actuated 2-fingered gripper that mimics a parallel jaw gripper's actions for this paper, but the method is applicable to a wide range of gripper morphologies.

After an algorithm predicts a grasp and a robot arm moves the gripper to the desired pose, we read data from each TOF sensor (with the gripper open). Each TOF sensor returns distance measurements from an 8x8 grid (64 zones) in a 90 degree diagonal field-of-view, which provides robust information about object geometry and position relative to the ends of the fingers (see Figure~\ref{fig:tof_visual}).

One reading from each sensor is passed to the classifier, which predicts grasp success or failure. Notably, this process occurs with the gripper in a fixed, open position \emph{where no contact is made with the object.} 

The reading and classification process cycles at 15 Hz, supporting deployment on grasping pipelines within real systems.

\subsection{Grasping Pipeline}\label{pipeline}
The typical grasping pipeline for unstructured environments is as follows:
\begin{enumerate}
    \item Capture visual and/or depth data of the scene.
    \item Identify target object(s)
    \item Generate grasp candidates
    \item Choose best grasp pose 
    \item Move gripper to target pose
    \item Attempt grasp
    \item If failed, go back to 1 and repeat
\end{enumerate}

This approach is limited as it requires grasp execution to determine failure. In cases of failure, the object may have been disturbed or displaced. As such, the process must start from the beginning with imaging the scene, and then predicting grasps again, which is a time consuming process.

Our grasping classifier could be directly integrated into grasping pipelines and remove the requirement to start over. For example, the grasping process proceeds as usual, but stops before step 6. Here, the predictive action explained in Section~\ref{predict} is performed. If a successful grasp is predicted, the grasp is executed. If a grasp failure is predicted, return to step 4 and move to the next predicted grasp pose. This modified grasping pipeline is shown in Figure~\ref{fig:flowchart}.




\begin{figure*}
    \centering
    \includegraphics[width=0.9\textwidth]{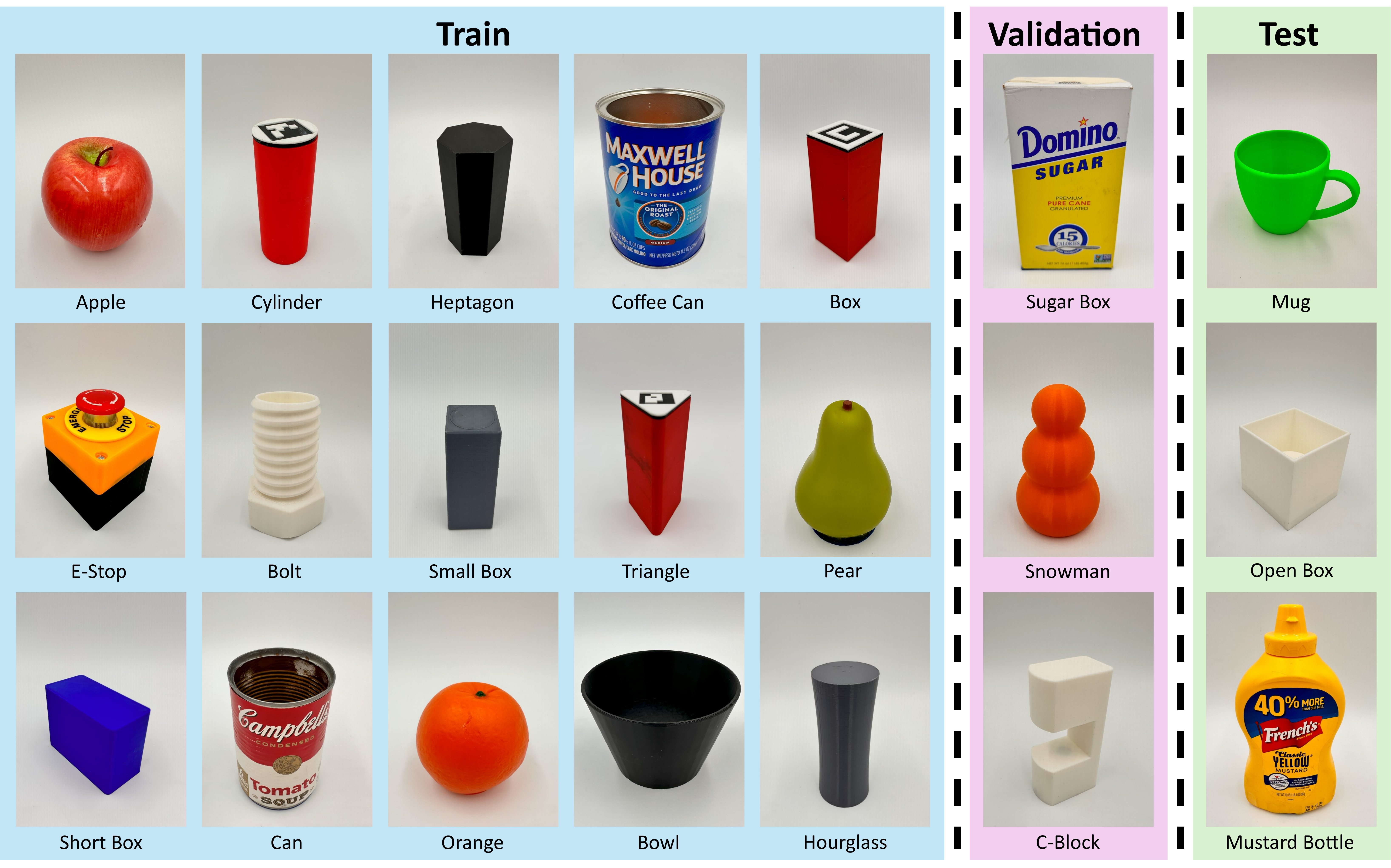}
    \caption{The objects used within this paper. Left --- Objects used for classifier training. Center --- Objects used for classifier validation and selection. Right --- Objects for final classifier testing.}
    \label{fig:object}
\end{figure*}
\subsection{Training and Classification}\label{train}
\begin{table}
\caption{The six degree-of-freedom range of pose modifications for the box. These ranges sample the entire space around the object, capturing both successful and failed grasps.}
\begin{center}
\def\arraystretch{1.4}%
\begin{tabular}{c | c | c} 
  Axis &Min & Max \\ 
  \hline
  X-translation & -2 cm & 2 cm\\ 
  Y-translation & -2 cm & 2 cm \\ 
  Z-translation & -2 cm & 2 cm \\
  Roll & -35 degrees & 35 degrees \\
  Pitch & -35 degrees & 0 degrees \\
  Yaw & \multicolumn{2}{c}{0, 15, 30, and 45 degrees} \\
\end{tabular}
\end{center}
\label{table:poses}
\end{table}

    
In order to train a machine learning classifier to predict stability, we collected over 2,500 grasps across 15 objects as shown in Figure~\ref{fig:object}. The objects were a combination of geometric shapes and objects from the YCB object set that were modified for use with the Grasp Reset Mechanism (see Section~\ref{hardware2}). For network validation and selection during training, we captured an additional 120 trials with three unseen objects. Another 120 trials were captured with an additional three unseen objects to test accuracy on our selected network after training. All trials and data were captured with real-world grasp attempts --- no simulated data was used.

\subsubsection{Data Collection}
For a diverse range of grasp poses for training, we use discrete uniform sampling of the gripper's location with respect to the object. The Grasp Reset Mechanism supports repeatably returning an object to the same position on the table, with any desired planar orientation. For non-symmetrical objects we placed the object in four different planar orientations (yaw). The grippers's location was varied in and out ($x$), left and right ($y$), up and down ($z$), and rotate up-down to change pitch ($p$) and around the wrist axis ($r$). The allowable amount of offset was set per object to generate an approximate 50/50 split of success and failure. An example of the range of poses for the box are shown in Table~\ref{table:poses}.

For seven base objects (cylinder, box, small box, triangle, pear, can, hourglass), over 300 trials were collected. The remaining eight training objects had 40 trials each randomly sampled out of the complete pool of discrete poses. The three validation and three test objects also each had 40 trials randomly sampled out of the complete pool of discrete poses. To reiterate, the pose sampling was chosen to create a roughly 50/50 split of successful and unsuccessful grasps.

For each trial, the arm moves the gripper to the grasp location. Then, the recording from the time-of-flight sensors begins at 10 Hz. After one second the gripper is closed (with joint angles recorded) and finally data collection is stopped. Subsequently, the arm moves the gripper to a drop off point 20 cm up and 10 cm away from grasp location and drops the object. If the object is grasped and remains grasped until dropped at the drop off point, the trial is marked as a success. The Grasp Reset Mechanism then returns the object to its starting location and orients it.

For this work, we assumed all objects are graspable by the robotic gripper. All objects are rigid or semi-rigid (the pear, apple, orange, mustard, and sugar box). Object masses range from 35.5 grams for the apple to 161.8 grams for the e-stop. No object properties, including mass or friction, are included in the classifier data.

\subsubsection{Data cleaning and reduction}
\label{sec:data_cleaning}

We first pre-processed the data by removing all unnecessary data from the time-of-flight target readings (the time-of-flight sensor reports multiple targets per zone). We found that this additional data was largely empty, contained no valuable information, and removing it eliminated 1,920 features. Additionally, we capped all distance readings at 12.5 cm. This removes any influence from background objects as it only considers points within the hand workspace.

After reduction, this leaves 1,028 features. This includes finger joint angles, as well as eight readings for each zone (8x8, 64 zones per sensor) on each of the two sensors. The seven readings include distance, standard deviation, reflectance, signal amount, number of targets, number of spad arrays activated, and reading quality.

\subsubsection{Classification}

We evaluated both random forests and neural networks as possible classification models. Brief testing showed both had similar performance, but neural networks were significantly more costly in terms of computation and time. This mirrors a similar comparison in previous work on grasp prediction~\cite{nigel}.  As a result, this work only explores random forest classification.

For our random forest classifiers, we optimized multiple parameters:
\begin{itemize}
    \item Number of trees
    \item Minimum number of samples to split a node
    \item Maximum tree depth
\end{itemize}

The final network was chosen as having the highest area under the curve (AUC) value on the unseen validation objects. In Section~\ref{results} we present the final network and compare results. 

To evaluate if ``more'' data is better, we also trained with data from two finger positions: the starting pose plus a second pose when the fingers had started to close (after 0.5 seconds). This did not significantly improve the classification results; for this reason, we did not pursue a more complicated classifier approach such as a LSTM.

\subsection{Hardware}\label{hardware2}
This work uses three hardware components, a custom gripper as well as the Grasp Reset Mechanism~\cite{dufrene2024grasp} paired with a Kinova Gen 3 7-DOF arm. 

\subsubsection{The Gripper}
The gripper used in this work is a custom two-finger, fully-actuated gripper. The physical design of the fingers is adapted from previous work by Nave et. al~\cite{kandk}. Each finger has two links, with each link independently actuated by a Dynamixel XL-330 servo motor. The distal links have a layer of silicone for an increased coefficient of friction. Within each distal link is a multi-zone time-of-flight sensor, a ST Microelectronics VL53L7CX (see Figure~\ref{fig:tof_visual}) \cite{tof}. Additionally, while not used in this work, IMUs are embedded in the distal links for additional sensing capabilities. Each of the four sensors (two TOF, two IMU) are connected to the Raspberry Pi on independent I2C buses. 

For each TOF sensor, a small opening in the silicone fingerpad allows unobstructed sensing perpendicular to the distal link. Each multi-zone TOF sensor returns distances measurements from 64 zones, an 8x8 grid, in a 90 degree diagonal field-of-view. The TOF sensors additionally return other sensor metadata, such as other targets within each zone, ambient light, reading confidence, and more. Including this additional metadata did not improve classification results, so we only used the 64 distance measurements.

All electrical and control components are located in the palm of the gripper. This includes a Raspberry Pi 4b, Dynamixel U2D2 and power hub, and 24V to 5V buck converter. The only requirements to use the gripper are a 24v power and WiFi (or Ethernet). These can be supplied to the gripper via the tool interface on some robotic arms, or provided through external cable routing. 

The gripper's software is implemented in the Robot Operating System 2 (ROS 2). Nodes read and publish data from each of the sensors, and a custom action server controls finger position with the Dynamixel servos. 

Although the gripper is fully actuated, we used a simple controller that mimics a parallel jaw gripper for the grasp. 


\section{Results}\label{results}
\begin{figure}
    \centering
    \includegraphics[width=0.49\textwidth]{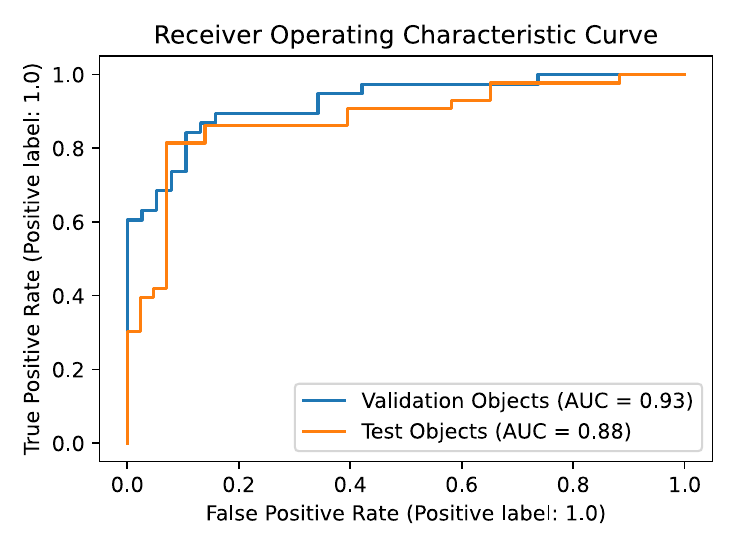}
    \caption{ROC curve comparing unseen validation objects (used for model selection) with unseen test objects.}
    \label{fig:roc}
\end{figure}
\begin{figure*}
    \centering
    \includegraphics[width=0.99\textwidth]{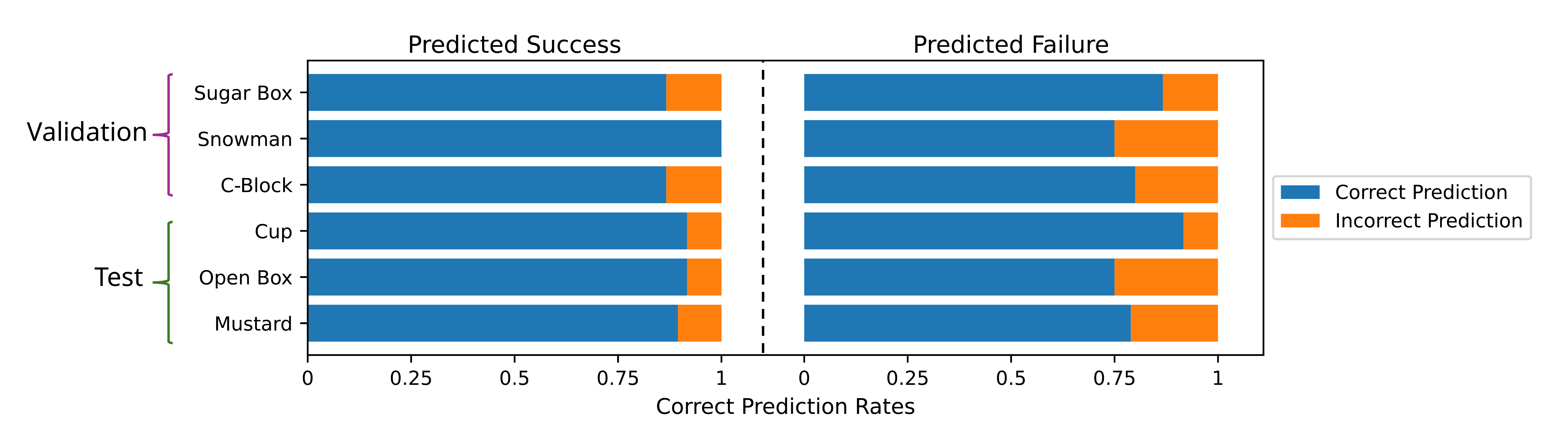}
    \caption{A bar chart visualizing a confusion matrix on a per-object level (three unseen validation and test objects). The left column shows true positive (blue) and false positive (orange) rates for grasps predicted as success. The right column shows true negative (blue) and false negative (orange) rates for grasps predicted as failures. These results are from the best-performing classifier with a decision boundary of 0.6.}
    \label{fig:bar}
\end{figure*}







In this section we present results from data collection (Section~\ref{data_cp}) and model training and classification (Section~\ref{model_train}). Overall, we demonstrate an accuracy of 86.0\% on the unseen test objects.

\subsection{Data Collection and Processing}\label{data_cp}
Over 3,000 real trials were conducted across the 15 training, 3 validation, and 3 test objects. The initial data set was fairly well-balanced between success and failure, although select objects had significant imbalance (the pear was particularly difficult to grasp and had a split of 90\% failures to 10\% successes). To create an equal distribution of successful and failed trials for each object, trials were randomly removed until each object had a 50-50 split. As a result, the {\em training} object dataset consists of 1,990 trials. For all model training discussed, the training object set was split 80/20, with 80\% of trials used for model training and 20\% used for \textit{seen} object validation. The unseen validation object set (sugar box, snowman, and c-block) includes 76 trials and the unseen test object set (mug, open box, and mustard bottle) includes 86 trials. 

As mentioned in Sections~\ref{hardware2} and~\ref{sec:data_cleaning}, the TOF sensors return more than just the 8x8 distance readings (each zone returns a 2nd, 3rd, and 4th target which largely were zero). We compared model performance for using just 1, 2, etc of these additional target readings. We validated that maximum model accuracy and area under the ROC curve (AUC) remained essentially identical without the added features. 

\subsection{Model Training}\label{model_train}
The AUC for the best model was 0.925, and accuracy on training objects 85.7\%, unseen validation objects 85.5\%, and unseen test objects 86.0\%. Note, these accuracies are with a decision boundary of 0.6, as false negatives are preferable compared to false positives. This best performing random forest had 60 trees, a minimum number of 5 samples to split a node and a maximum depth of 16.
\begin{table}[h]
\caption{Confusion matrix for the three unseen \textit{validation objects} over 76 trials. The values are with a decision threshold of 0.6.}
\begin{center}
\def\arraystretch{1.4}%
\begin{tabular}{m{7.25em} | m{8em} m{2cm}} 
  &\textbf{Successful Grasp} & \textbf{Failed Grasp} \\ 
  \hline
  Predicted Success & 89.5\% & 10.5\%\\ 
  Predicted Failure & 18.4\% & 81.6\% \\ 
\end{tabular}
\end{center}
\label{table:cm_1}
\end{table}

\begin{table}[h]
\caption{Confusion matrix for the three unseen \textit{test objects} over 86 trials. The values are with a decision threshold of 0.6.}
\begin{center}
\def\arraystretch{1.4}%
\begin{tabular}{ m{7.25em} | m{8em} m{2cm}} 
  &\textbf{Successful Grasp} & \textbf{Failed Grasp} \\ 
  \hline
  Predicted Success & 90.7\% & 9.3\%\\ 
 Predicted Failure & 18.6\% & 81.4\% \\ 

\end{tabular}
\end{center}

\label{table:cm_2}
\end{table}



Receiver operating characteristic curves (ROCs) for this model on the unseen validation objects and unseen test objects are shown in Figure~\ref{fig:roc}. The similar AUC values between unseen validation and test objects, while slightly lower on the test objects, demonstrates a well-fit model with minimal over-fitting. 

Confusion matrices for both the unseen validation data and unseen test data are presented in Tables~\ref{table:cm_1}~and~\ref{table:cm_2}. Like above, values are reported with a decision boundary of 0.6 to reduce the rate of false positives. With this decision boundary, our model showed strong success prediction performance with approximately 90\% true positive predictions and 10\% false positive predictions across both the validation and test objects.

When evaluated on individual objects (validation and test), correct prediction rates vary from 86.6\% to 100.0\% for trials predicted as a success and 75.0\% to 91.7\% for trials predicated as a failure. This is shown in detail in Figure~\ref{fig:bar}. Again, a decision boundary of 0.6 was used.

We additionally tested taking two sensor readings, with the second reading occurring after the fingers started closing. With the additional readings, maximum AUC was 0.925 and accuracy was 83.7\% on train objects and 86.8\% on validation objects. Although accuracy on validation objects improved slightly (by 1.3\%), AUC was the same and training accuracy decreased by 2\%. We determined that multiple sensor readings provide negligible benefit.



\section{Discussion}
\begin{figure}
    \centering
    \includegraphics[width=0.49\textwidth]{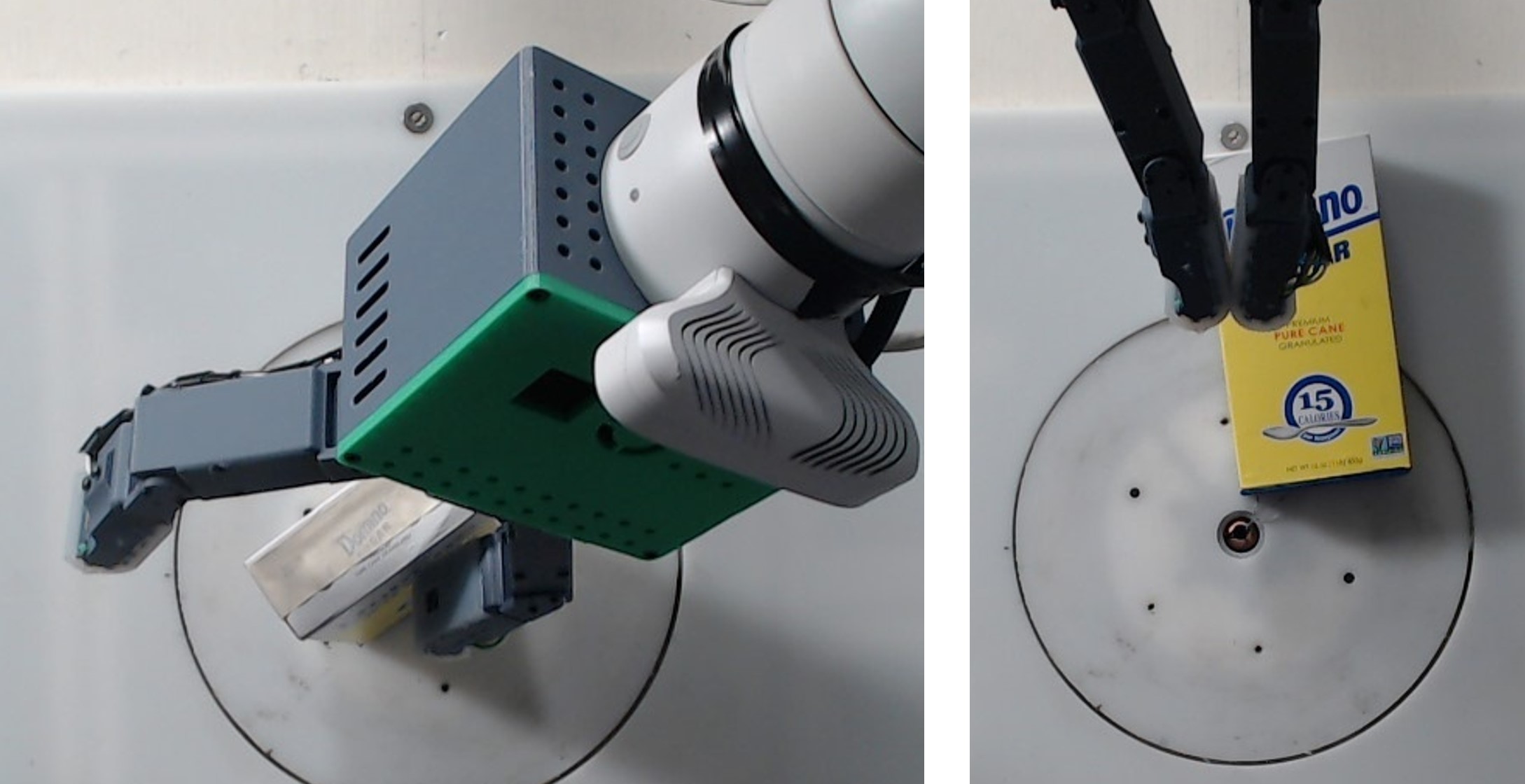}
    \caption{A trial with the sugar box which failed, but was incorrectly predicted as a success by the trained model. The left image is the gripper at the grasp pose before it closes on the object. The right picture shows the gripper after it moved to the drop off location before opening.}
    \label{fig:failed_sugar}
\end{figure}
We demonstrated strong grasp stability prediction accuracy with our time-of-flight approach, with 86.0\% accuracy on test objects. We identified several areas to be addressed in future work to improve this accuracy. For example, our model has no information about object weight, size, surface texture/friction, or rigidity of the target object. Tactile approaches may gain some information after grasping, such as rigidity and object weight. Our contact free prediction would require external sources for that information.

The importance of this information is visible in a trial with the sugar box (see Figure~\ref{fig:failed_sugar}). The grasp is located at the top corner of the sugar box, and is not perpendicular. This grasp was predicted to succeed by the model, and likely would have given a similar pose relative to the standard YCB box. However, in the case of the sugar box the grasp was far off center from the center of mass, leading the sugar box to rotate and then drop out of the grasp. Additionally, the surface of the sugar box is slicker (lower friction coefficient) than the standard box. This trial is shown in . As a result, incorporating additional information such as estimated object weight and size from an outside grasp planner may improve accuracy, which has been successful for tactile approaches~\cite{multimodal}. 

Another area of further exploration is the impact of gripper design and actuation. Within this work, we used a fully actuated hand mimicking the behavior of a simple parallel jaw gripper. Grasp closure and force was inherently not as precise as a rigid parallel jaw gripper. We propose further testing our method with other grippers, as more repeatable grippers likely will improve classification accuracy. 

Many current grasp planners, such as GPD~\cite{gpd}, use point clouds to generate grasp candidates. Our method with multi-zone time-of-flight sensors could additionally compliment point cloud based planners. Information from the point cloud can be combined with time-of-flight readings to provide more information to the grasp stability classifier. Or, the time-of-flight readings could be used to improve/fill in gaps in the initial point cloud for more accurate grasp prediction.

\section{Conclusion}
In this work we presented an approach to grasp stability classification using in-hand time-of-flight sensors. Our approach relies solely on time-of-flight sensor data and joint angles, and does not require grasping or contacting the object. After collecting more than 2,500 real trials, we trained a random forest classifier and demonstrated 86.0\% accuracy on the test set of three unseen objects. Similar results across the training, validation, and test objects demonstrate the generalizability of our model to other objects. 

Our results demonstrated that a contact-free approach can be used for grasp stability classification and to enhance existing grasping pipelines. We also discussed further areas for improvement, including providing the classifier with additional information, such as the size and mass of an object. 

Beyond classification, we believe these results show that in-hand TOF sensors are a useful modality in grasping and manipulation. Future work could explore active servoing to optimal grasp locations, using TOF sensors for object or scene reconstruction, or using the sensors for more complex manipulation tasks.


\bibliographystyle{IEEEtran}
\bibliography{main}
\end{document}